\documentclass[runningheads]{llncs}
\usepackage[T1]{fontenc}
\usepackage{graphicx}
\graphicspath{{figs/}}
\usepackage{subcaption}
\usepackage{marvosym}
\usepackage{booktabs}

\begin{document}
\title{Reading the Table: Sim-to-Real Game State Detection for the Chef's Hat Card Game}
\titlerunning{Reading the Table: Sim-to-Real Game State Detection}
\author{Connor Gäde\textsuperscript{(\scalebox{1.2}{\Letter})}%
\and
Matthias Kerzel%
\and
Hauke Bünning%
\and
Jan-Gerrit Habekost%
\and
Mariana Barros%
\and
Stefan Wermter%
}
\authorrunning{Gäde et al.}
\institute{Knowledge Technology, Dept. of Informatics, University of Hamburg, Germany\\
\email{\{connor.gaede, matthias.kerzel, stefan.wermter\}@uni-hamburg.de}\\
\url{www.knowledge-technology.info}
}
\maketitle              %
\begin{abstract}

Tabletop games offer a compelling scenario to study the interaction between artificial and human agents, combining social interaction with game state tracking and physical manipulation. To assess the game state, a robot needs to precisely detect closely related game objects, such as different card designs, requiring dedicated training data that is costly to annotate in the real world. Instead, we develop a fully automated synthetic data collection pipeline, using re-projection of known 3D coordinates to generate labels automatically and domain randomization to bridge the reality gap. This approach is used to collect a dataset of 25413 samples to finetune an object detection and a segmentation model, which are combined to assign detected cards to game-relevant zones and infer the game state. When evaluated on a real NICO humanoid robot, our pipeline achieves an F1-score of 0.906 for card-to-zone assignment, thus providing the perceptual foundation for a robot that can autonomously play Chef's Hat against human players.

\end{abstract}
\section{Introduction}

\begin{figure}[tp]
\centering
\begin{subfigure}[t]{0.48\textwidth}
  \centering
  \includegraphics[width=\linewidth]{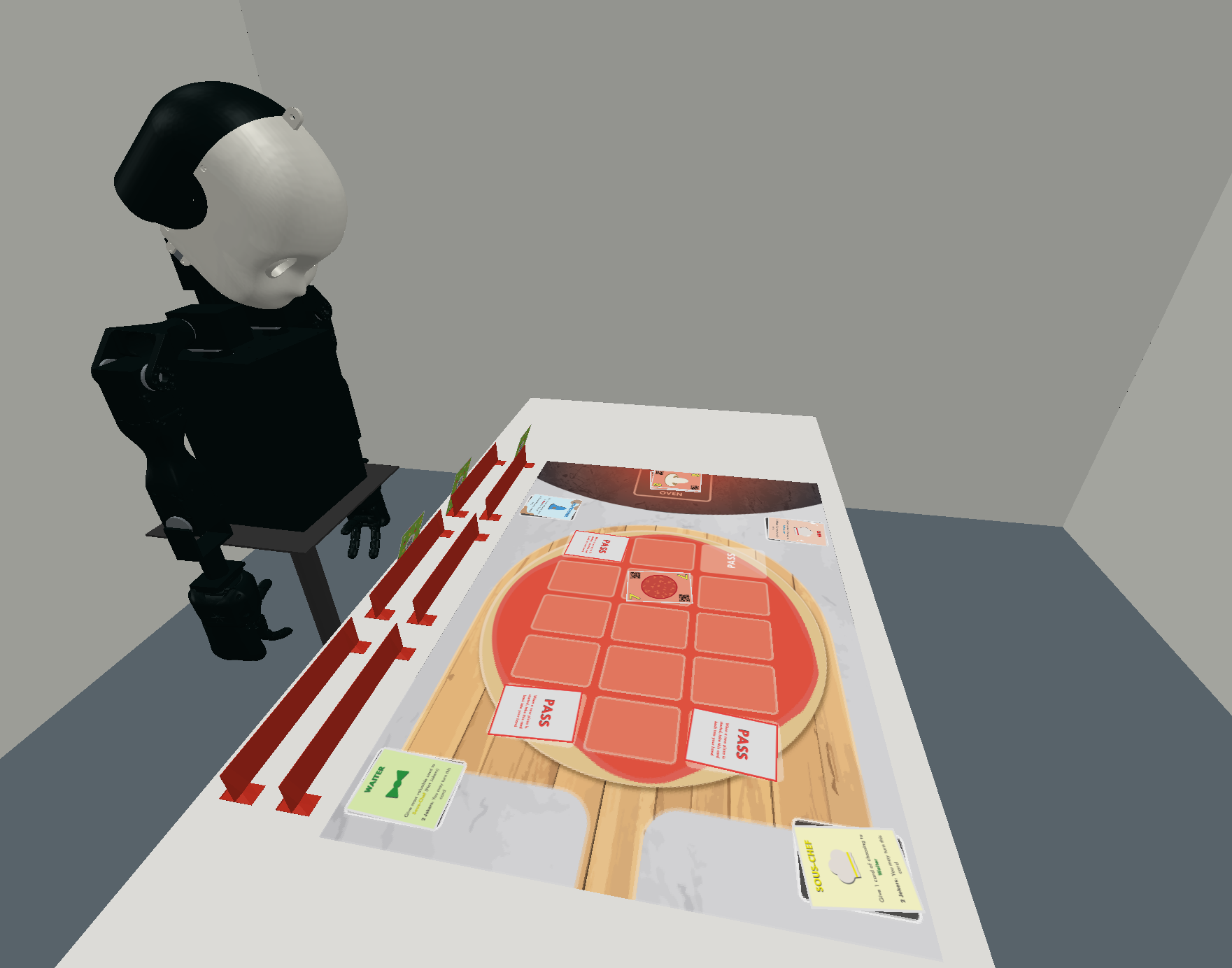}
  \caption{}
  \label{fig:nico_sim}
\end{subfigure}\hfill
\begin{subfigure}[t]{0.5025\textwidth}
  \centering
  \includegraphics[width=\linewidth]{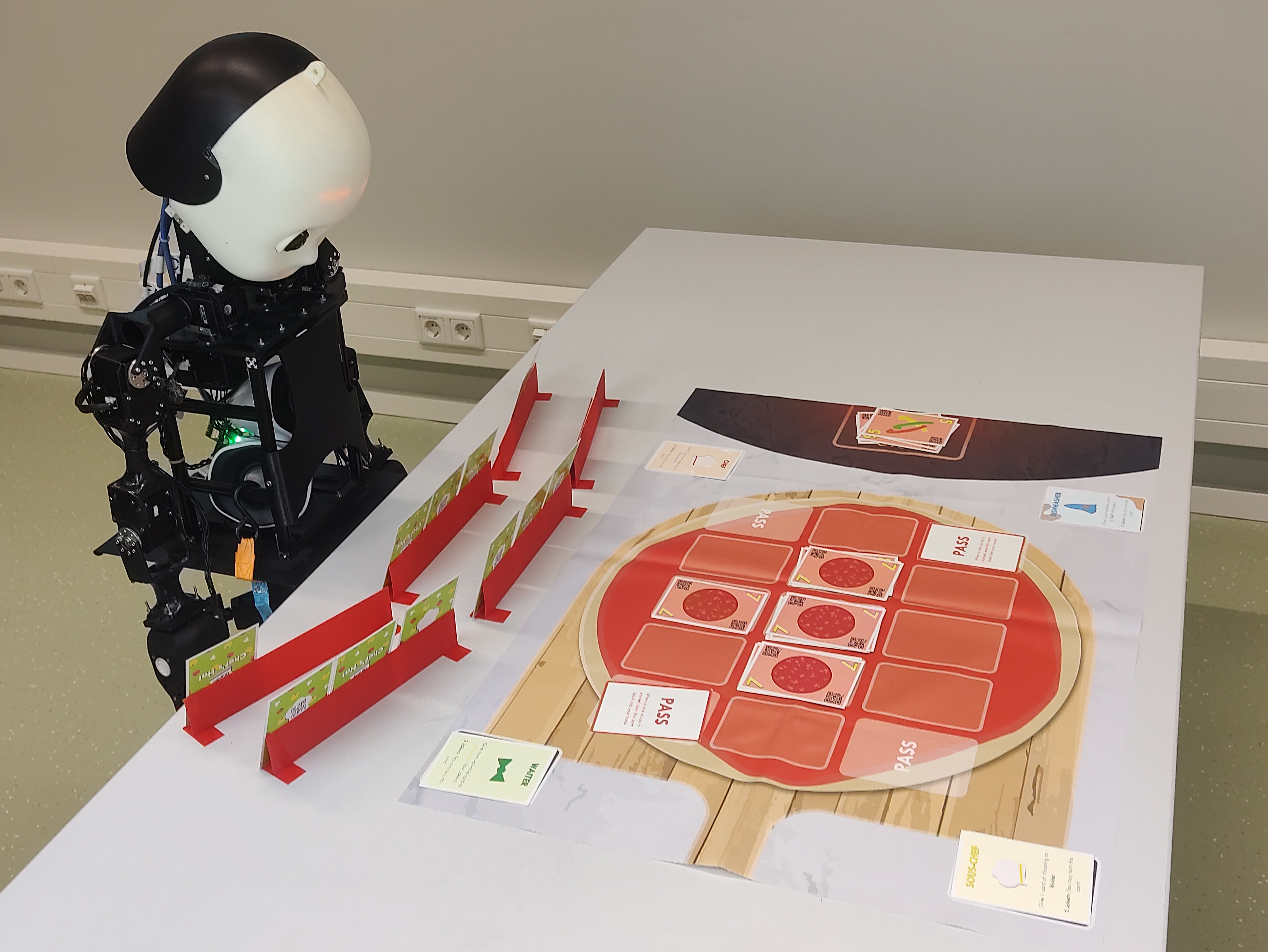}
  \caption{}
  \label{fig:nico_real}
\end{subfigure}
\caption{Simulation (a) and real-world (b) setup}
\label{fig:nico_setup}
\end{figure}

Enabling robots to participate in multiplayer tabletop games with human players combines multiple challenges in robotics and AI. The robot needs to naturally interact with the other players, accurately infer and track the game state, and perform fine-grained object manipulation to execute game actions. Recent advances in embodied robot agents utilize an LLM \cite{allgeuer2024robots} or VLM \cite{gade2024embodying} for high-level interaction, scene understanding, and task-planning, combined with detection and action modules for object localization and manipulation. While these agents generally use open-vocabulary detectors to interact with everyday items \cite{allgeuer2024robots,gade2024embodying}, task domains with visually similar objects or specific class labels require specialized detection models and thus training data \cite{gajdosech_shaken_2025}.
A card game like Chef's Hat represents one such domain, where visually similar cards must be distinguished by value and their placement in specific playing field zones identified.

Collecting training data for such tasks on a real robot requires meticulously annotating bounding boxes for each card and segmentation masks of the playing field by hand, which is time-intensive, prone to errors, and can introduce biases towards specific camera poses, lighting conditions, or object placements, making a trained model generalize poorly. Simulators provide a more scalable alternative, allowing the free arrangement of objects in the scene and automatically generating labels through re-projection using the known object poses and camera parameters. However, due to the \emph{reality gap} caused by lower visual fidelity and simplified assumptions of a simulator compared to reality, models trained on synthetic data generalize poorly to real-world input. Domain randomization \cite{tobin_domain_2017} mitigates this gap by altering visual and environmental parameters of the simulation to increase the resilience of a trained model to domain shifts and thus its ability to interpret the unseen real-world domain.

In this work, we develop a fully automated synthetic data collection pipeline in CoppeliaSim\footnote{\url{https://www.coppeliarobotics.com}} and create a game state detection approach for the Chef's Hat card game to play it on a real humanoid robot. For this, we create a dataset of 25413 samples consisting of paired domain-randomized and non-randomized canonical images showing the Chef's Hat playing cards on the playing field and custom 3D-printed cardholders. %
Each image is labeled with card identities, zone assignments, and the projected 2D image coordinates for individual cards and playing field zones, from which bounding boxes and segmentation masks can be derived. Each sample further includes the 6D simulation poses of all objects, camera parameters, and a random seed, allowing the reconstruction of any scene in simulation and enabling future extensions such as card pose estimation or motion planning for physical manipulation.
With this synthetic dataset, we finetune an object detection model to detect playing cards and a segmentation model to determine different zones of the playing field, combining them to determine which cards are placed in which zone. When evaluated on real-world data without additional finetuning, our model assigns the correct cards to the correct zones with an F1-score of 90.6\%, enabling valid game actions in most situations.

\section{Related Work}

\subsection{The Chef's Hat Card Game}
Chef's Hat is a four-player tabletop game played with cards, a game map, and optional props like a chef's hat or a dishwasher's towel to indicate different roles. The game is a competitive card-shedding game, where players try to discard their hand. The first player to succeed is crowned the new chef, while the later places go to the sous-chef, the waiter, and finally the dishwasher. These roles are randomly distributed in the beginning and give certain players a slight advantage to further foster competition among the players for the first places. However, the game is also fast-paced, and players switch roles quickly. From the human-centered perspective, the game is designed to be socially engaging, requiring players to reason about what cards other players might have and what they might do while engaging in friendly banter. For an artificial agent, there is the added challenge of tracking and later also manipulating playing cards.
While an artificial agent is taking part in a physical game, interacting with the other players and playing the game on a level that is engaging for fellow human players is no small challenge; all aspects of the game are designed to be \emph{within arm's length} of the state-of-the-art. Reinforcement learning \cite{barros2021you} and large language models \cite{pereira2024there} have been explored to learn how to compete; explainable AI is being used to allow the agent to communicate how they adapt to other player strategies. These models were, in part, informed by data collected from real players in a simulation environment\footnote{\url{https://github.com/pablovin/ChefsHatGYM}}. Now it is time to bring real robots to the table.

\subsection{Sim-to-Real Transfer}

Training a perception model that reliably detects playing cards or segments zones of the playing field on camera images of a real robot requires a sufficient amount of annotated, domain-specific training data. In a simulated environment, objects can be freely arranged in the scene and annotated by projecting known 3D coordinates into the camera frame, enabling a fully automated data recording and labeling process for synthetic datasets. However, models trained on synthetic data often perform poorly when deployed in a real environment due to the visual and physical differences between the simulated and real domain, the \emph{reality gap}.

The two main methods to reduce this gap are domain randomization and domain adaptation \cite{kleeberger2020survey,horvath2022object,gade_domain_2024}. While pixel-level domain adaptation \cite{bousmalis2017unsupervised} approaches attempt to shift the synthetic training images towards the real-world target domain, requiring at least a minimal amount of real-world reference data to train a separate adaptation model \cite{rao2020rl,ho2021retinagan}, domain randomization \cite{tobin_domain_2017} instead attempts to make the trained model itself more robust to shifts in the input domain by altering visual properties \cite{tobin_domain_2017,gade_sim--real_2022} or dynamics \cite{peng2018sim} of the simulation. Given a sufficient amount of randomization, the distribution of the synthetic data should theoretically cover the properties of the real domain and thus allow a model to directly interpret real-world images \cite{westerski2024synthetic}. Domain randomization is commonly used to generate synthetic data for object detection tasks \cite{tobin_domain_2017,westerski2024synthetic}, including ones with similar classes \cite{horvath2022object}, as well as robot control tasks \cite{james_transferring_2017,gade_sim--real_2022}. Other approaches attempt to combine domain randomization and domain adaptation by learning to transfer randomized synthetic data into a simplified canonical domain, thereby learning to simplify real-world data \cite{james2019sim,gade_domain_2024}. %
We opt for domain randomization over adaptation as it requires no real-world reference data and generalizes across environments and lighting conditions without additional training. As the fisheye distortion of the robot's eye cameras is difficult to model accurately in simulation, we specifically randomize the camera field of view.

\section{Methodology}

\subsection{Chef's Hat Cards and Playing Field}

The Chef's Hat game includes 68 ingredient cards with 13 unique front designs that are evenly distributed among players and have to be placed in zones on top of the pizza depicted on the playing field throughout the game. Additionally, each player is given a pass card that is placed in their designated pass zone of the playing field to indicate that the player will not place any more ingredients on the current pizza. When all but one player have passed, the ingredients on the pizza are moved to the oven zone of the playing field and a new pizza is started. Once all 68 ingredients are in the oven, a new round called a \emph{shift} is started, the ingredients are redistributed, and each player is awarded points and one of 4 unique role cards based on their finishing order in the last shift, which is placed in the 4 dedicated role zones of the playing field. %

\begin{figure}[tp]
    \centering
    \begin{subfigure}{0.49\textwidth}
        \includegraphics[width=\textwidth]{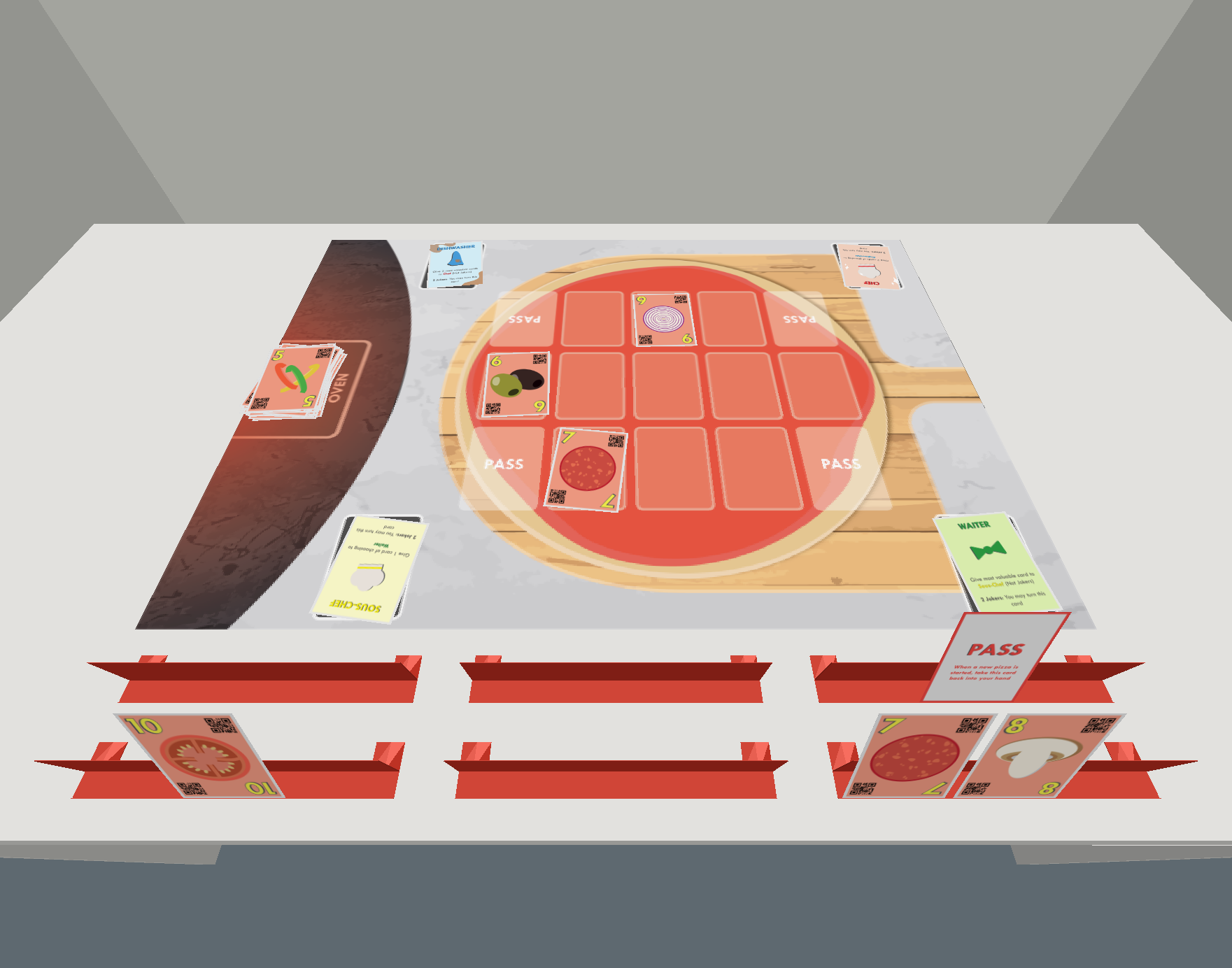}
        \caption{}
    \end{subfigure}
    \hfill
    \begin{subfigure}{0.49\textwidth}
        \includegraphics[width=\textwidth]{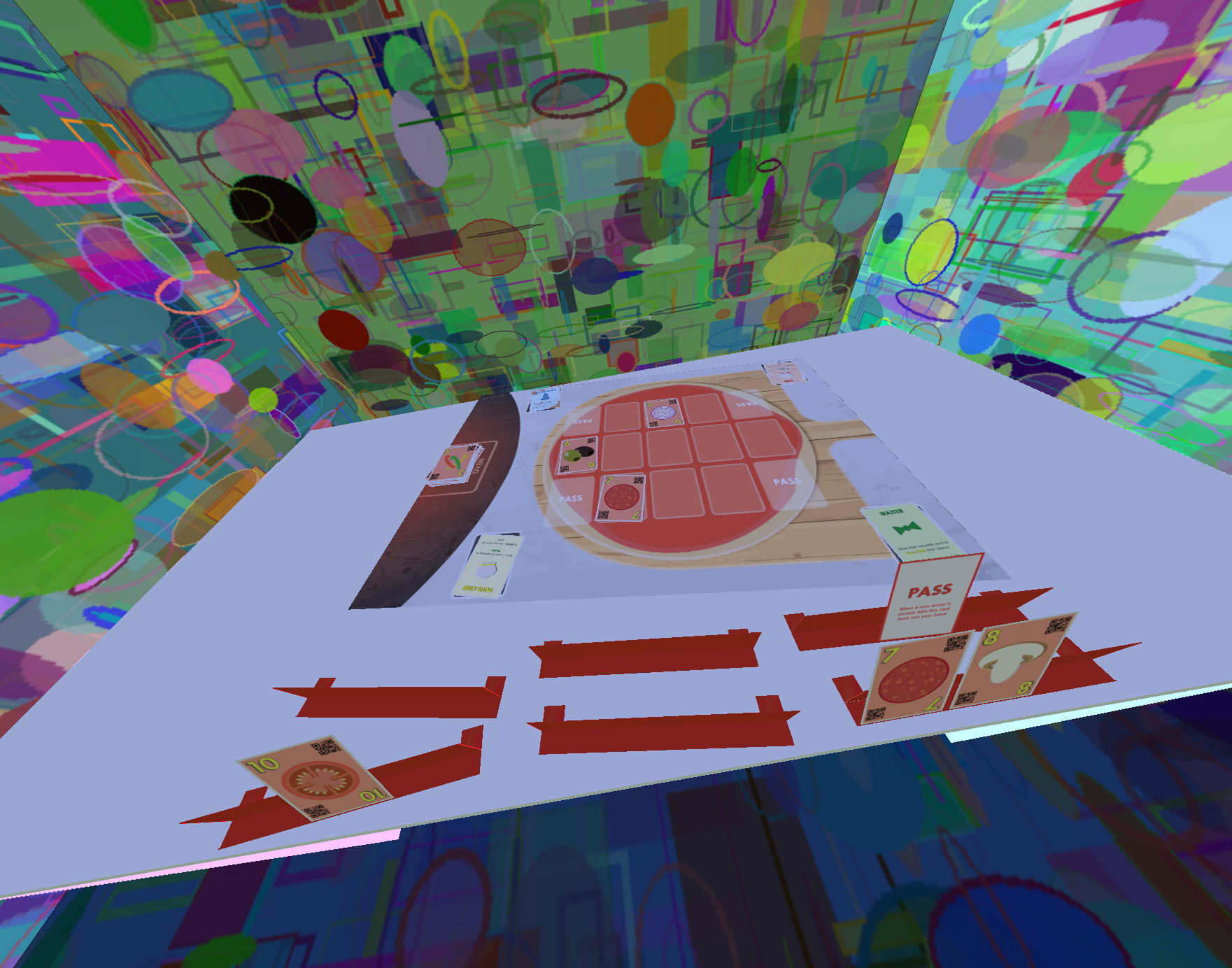}
        \caption{}
    \end{subfigure}
    
    \vspace{0.5em}
    
    \begin{subfigure}{0.49\textwidth}
        \includegraphics[width=\textwidth]{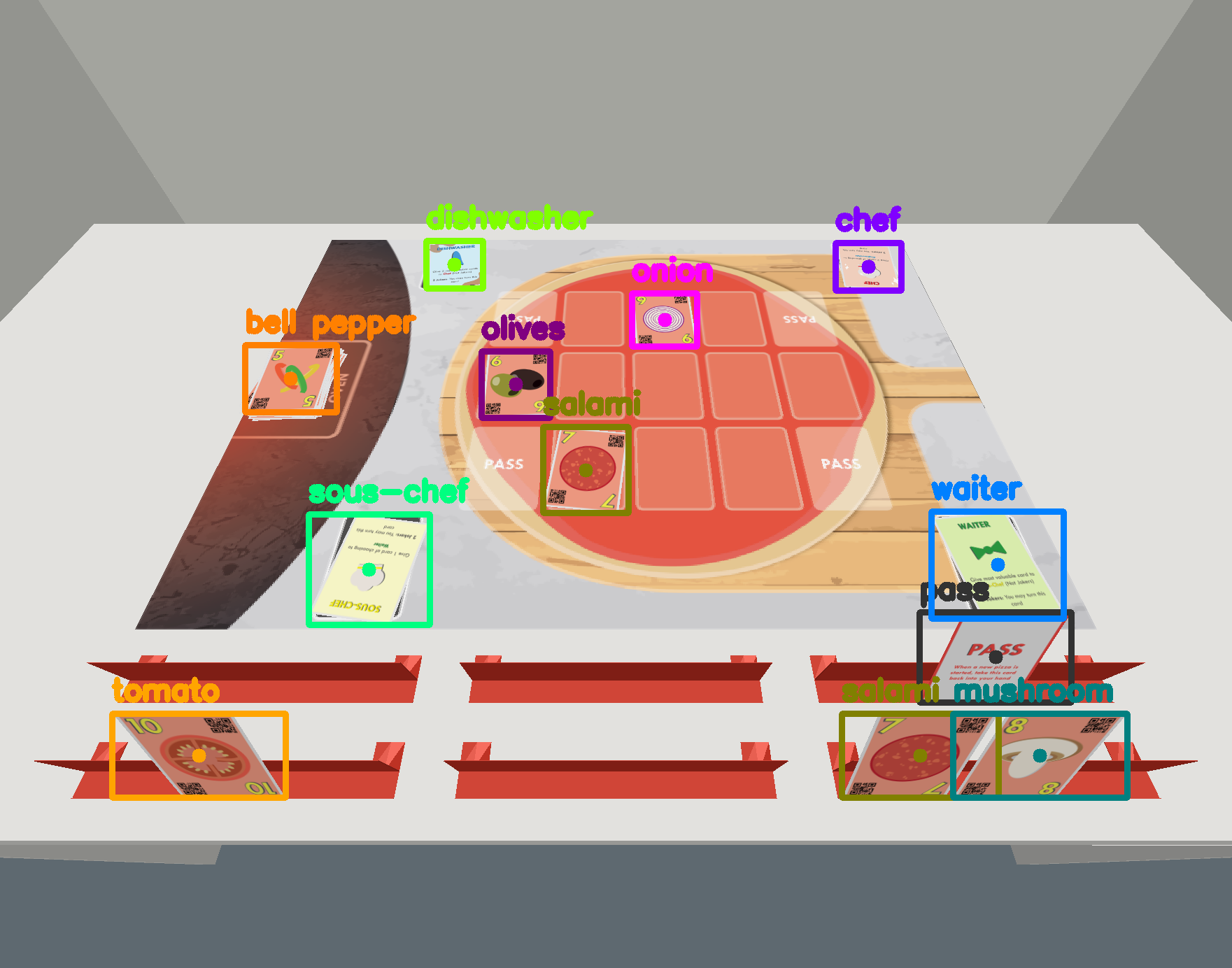}
        \caption{}
    \end{subfigure}
    \hfill
    \begin{subfigure}{0.49\textwidth}
        \includegraphics[width=\textwidth]{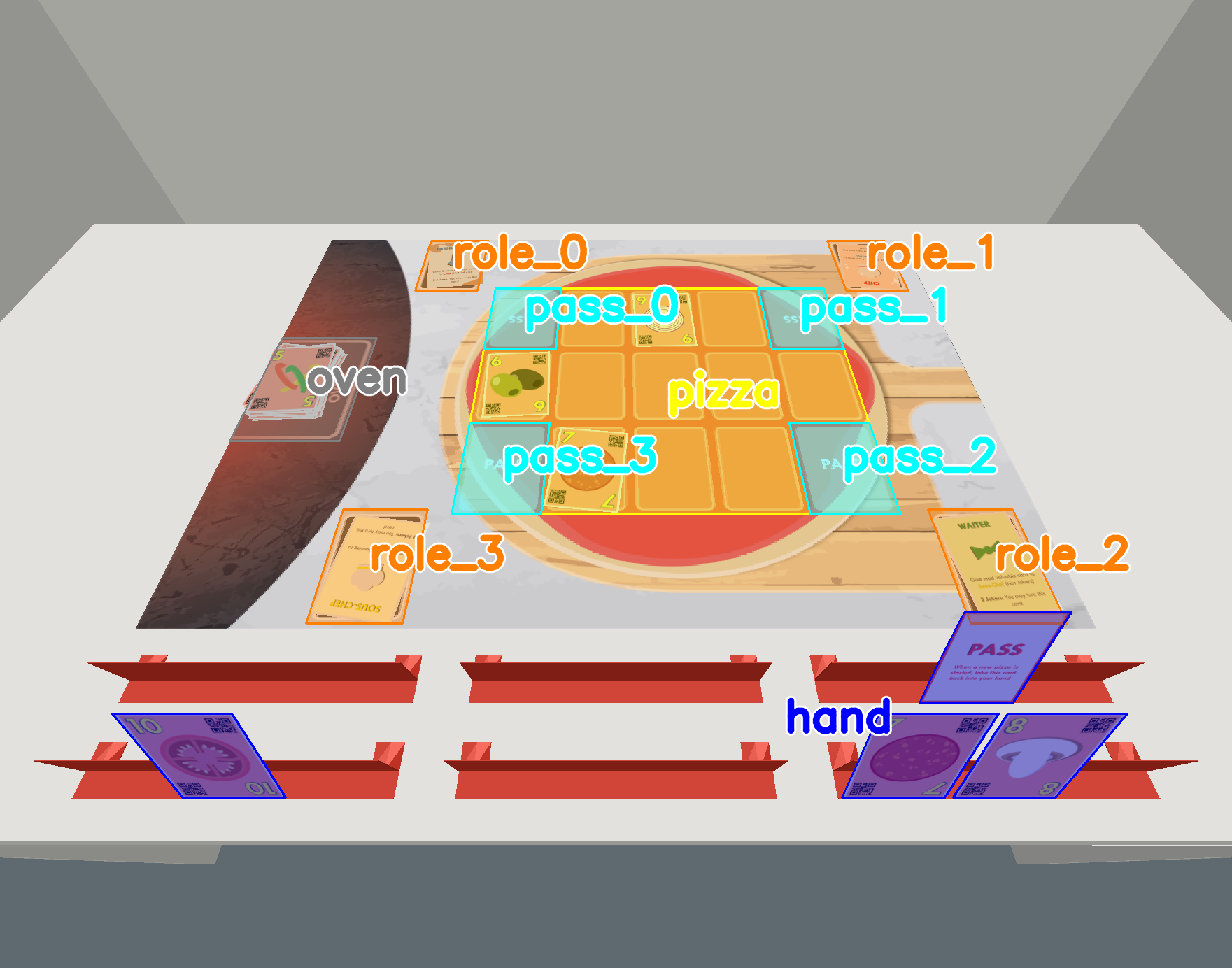}
        \caption{}
    \end{subfigure}
    \caption{Matching canonical (a) and randomized (b) simulation images, the corresponding ground-truth bounding boxes (c), and segmentation masks of the game-relevant zones (d).}
    \label{fig_sim_combined}
\end{figure}

Thus, the game requires detecting 18 unique card designs across 11 zones of the playing field (see Figure \ref{fig_sim_combined}), where the 13 ingredient cards share a largely identical layout, differing only in the central ingredient illustration and small value numbers, making fine-grained classification from a robot's perspective particularly challenging.
Although each ingredient card carries a unique QR code, these are too small to detect reliably under the fisheye distortion of the robot's cameras. We therefore restrict our system to rely solely on camera images without the help of additional markers, ensuring the robot processes the same visual information available to a human player. 

\subsection{Experimental Setup}

Our task is to infer the game state of the Chef's Hat game from the first-person view of a humanoid robot. To do so, we position the upper body of a NICO (Neuro-Inspired COmpanion) humanoid robot \cite{kerzel_nico_2017}, so that it faces a table on which the game is set up (see Figure \ref{fig:nico_real}). The robot is looking down towards the table at a 55° angle to record the area relevant to the game with the two fisheye cameras serving as its eyes. As the game requires each player to hold up to 18 cards at once, which is unfeasible with the child-sized hands of the robot, we utilize 3D-printed card stands to position the cards upright for the robot, so that it can see the front while other players can only see the back. Six of these card stands, each holding three cards, are placed at the edge of the table between the robot and the playing field, allowing the robot to observe both the cards it holds and those placed in the different zones of the playing field by itself and other players.

\subsection{Simulation Environment and Data Collection}

To automate the data collection process and create a sufficiently large dataset, we replicate our setup in CoppeliaSim, as seen in Figure \ref{fig:nico_sim}, and control the simulation via PyRep \cite{james_pyrep_2019}. For each square on the playing field that a card can be placed into, we attach an invisible dummy point to its center as a reference frame for the card placement. To account for small variations, each placed card is randomly moved by up to $2.5$ mm from its reference frame along both axes of the playing field and rotated by up to $0.15$ radians around the upwards axis. Additionally, the card is randomly flipped upside down by rotating it $\pi$ radians around the same axis. Multiple cards in the same zone get stacked on top of each other with an offset of $0.4$ mm. 

Similarly, each of the six card stands has three equally spaced dummy points as a reference frame to place cards into. As cards are staying upright on the card stand by design, cards placed here only get shifted by up to $2.5$ mm along one axis and randomly flipped upside down by $\pi$ radians without further rotation.

Additional dummy points are placed at the corners of each card and each square of the playing field. By projecting their 3D coordinates into the 2D image plane using the camera's known intrinsic matrix, we use these points to automatically define bounding boxes for each card and segmentation masks for the different playing field zones and cards on the card stands.

For our data collection procedure, we simulate games between multiple artificial agents. However, we simplify the rules for ingredient card placement in order to create a greater variety of game states and avoid implementing or running agents adhering to the exact game rules. The starting player is chosen at random. During each turn, we then randomly choose up to 11 cards for the current player that can be played. If that number is equal to or greater than the number of cards placed in the previous turn, those cards are placed on top of existing cards, and in random unoccupied zones if greater. Otherwise, that player plays their pass card. The rules for finishing pizzas, ending a shift, awarding points, and assigning roles remain the same, but we forgo any game actions associated with the roles. Once a player reaches 15 points, we fully reset the game and start a new round with no roles until 100 full games have been played.

Each new game state is recorded as a sample consisting of an RGB image, the camera's 6D pose and field of view, and for each card stand, card, and dummy point, as well as the playing field, its 6D pose along with its projected 2D image coordinates. Another sample of the same game state is then recorded after randomizing the scene without affecting card placement. This second sample also contains the seed of the random number generator for reproducibility. We then restore the default scene representation before proceeding to the next game state. Thereby, we create two matching variants of each board state, one maintaining a canonical representation as a baseline and another randomized representation to improve the sim-to-real transfer of a trained policy.

\subsubsection{Domain Randomization}

We visually alter the scene for our randomized samples by changing the camera perspective, lighting, and background textures of the environment, as well as the pose of the cardholders and the playing field.

Lighting and background textures are randomized to increase robustness to varying lighting conditions and background clutter across different setups and recording sessions.
All four light sources in the scene are moved up to 2 m in either direction along 2 axes at a fixed height of 2.5 m, oriented randomly, and have the color of both their specular and diffuse component randomized. For the floor and each wall, we generate a $512\times512$ px texture by randomly distributing 512 rectangles and 512 ellipsoids on top of a random base color. The shapes have a random width and height of up to 64 px, and a line width of up to 4px. Both the line color and the fill color are selected at random, with 50\% of the shapes consisting only of an outline without a fill color, 25\% having one solid color, and the remaining 25\% having different line and fill colors.
The camera pose and field of view are randomized to account for varying robot placements and the difficulty of accurately modeling the fisheye distortion of the robot's cameras in simulation, moving the camera up to 10 cm from its default position in either direction on all 3 axes, rotating it up to 0.2 radians in either direction around all 3 axes, and selecting a random field of view between 90° and 135°. Lastly, the playing field and cardholders are randomly moved and rotated on the table to reflect the changing, imprecise real-world placements. We randomly move the playing field by up to 10 cm to the left or right, up to 2 cm forward or backward, and rotate it up to 0.15 radians in either direction around the upward axis. Likewise, we move the cardholders up to 2 cm along the same axes and also rotate them by up to 0.15 radians.
As placed cards get the corresponding slot on the cardholder or playing field assigned as parent, they maintain their relative pose and move along with the parent when its pose is randomized. To ensure that all cards remain visible, we repeat the randomization process if the projected 2D image coordinate of any card center lies outside the image boundaries.

\subsection{Game State Detection Pipeline}

In order to infer the game state from the observed table, the robot needs to locate and classify all visible cards and assign each of them either to the held cards on the cardholders or to the zone of the playing field in which the card is placed. Thus, we divide our problem into two sub-tasks: Object detection to locate and classify the individual cards in view, and segmentation to identify the different game-relevant zones of the observation image that the cards have to be assigned to. To solve these two tasks, we finetune both the YOLO26n and YOLO26n-seg models from Ultralytics \cite{yolo26_ultralytics} for object detection and instance segmentation on our domain-randomized dataset, using the canonical data as the validation set. The models are trained for 25 epochs using the default training parameters. For each card detected by the object detection model, we take the center of the predicted bounding box and check whether it lies within the segmentation mask of one of the game-relevant zones predicted by the segmentation model, thereby enabling us to assign the individual cards to the game zones.

\section{Experiments}

\begin{figure}[tp]
\centering
\begin{subfigure}[t]{0.195\textwidth}
  \centering
  \includegraphics[width=\linewidth]{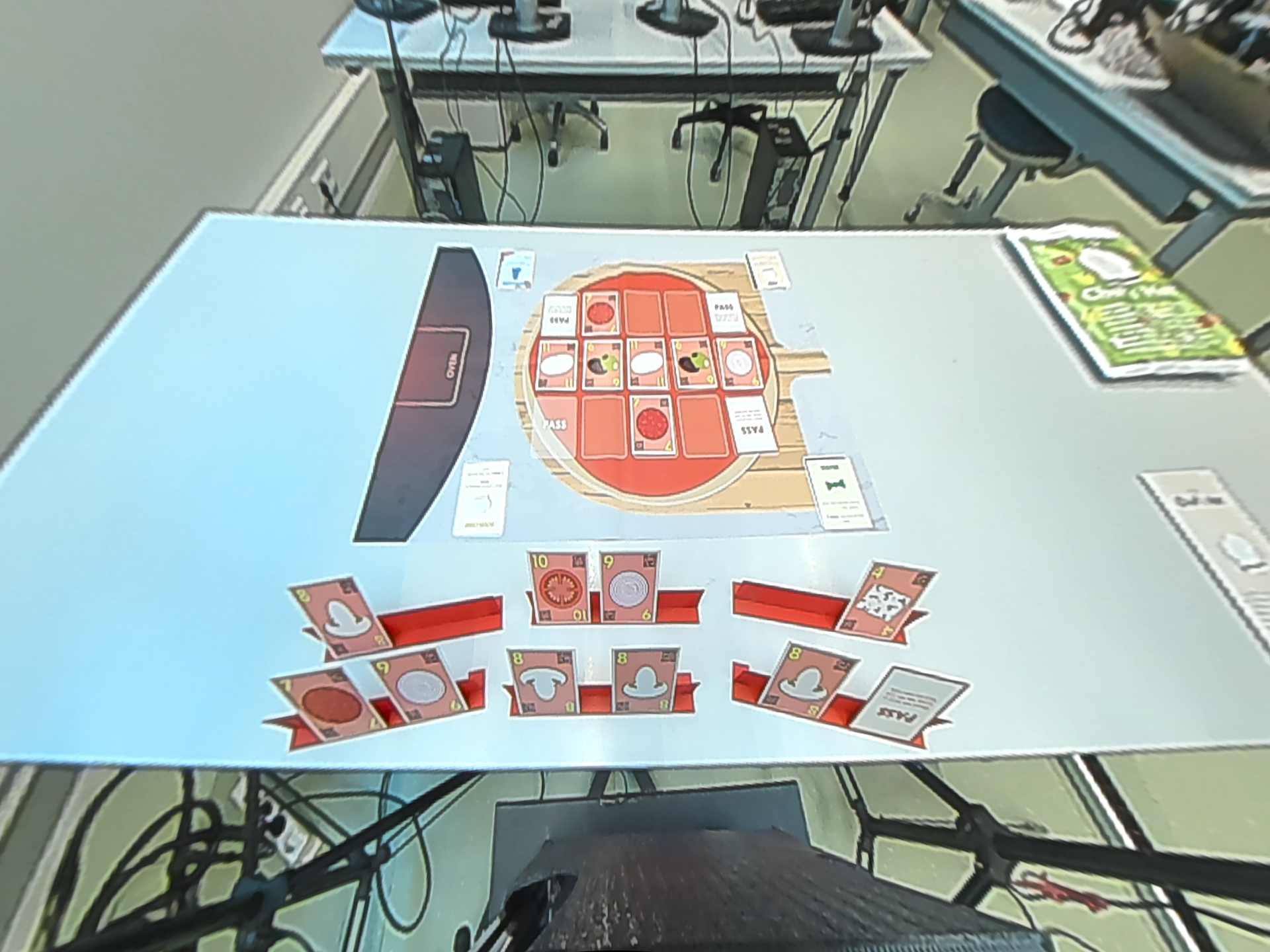}
  \caption{}
  \label{fig:real_data_random}
\end{subfigure}\hfill
\begin{subfigure}[t]{0.195\textwidth}
  \centering
  \includegraphics[width=\linewidth]{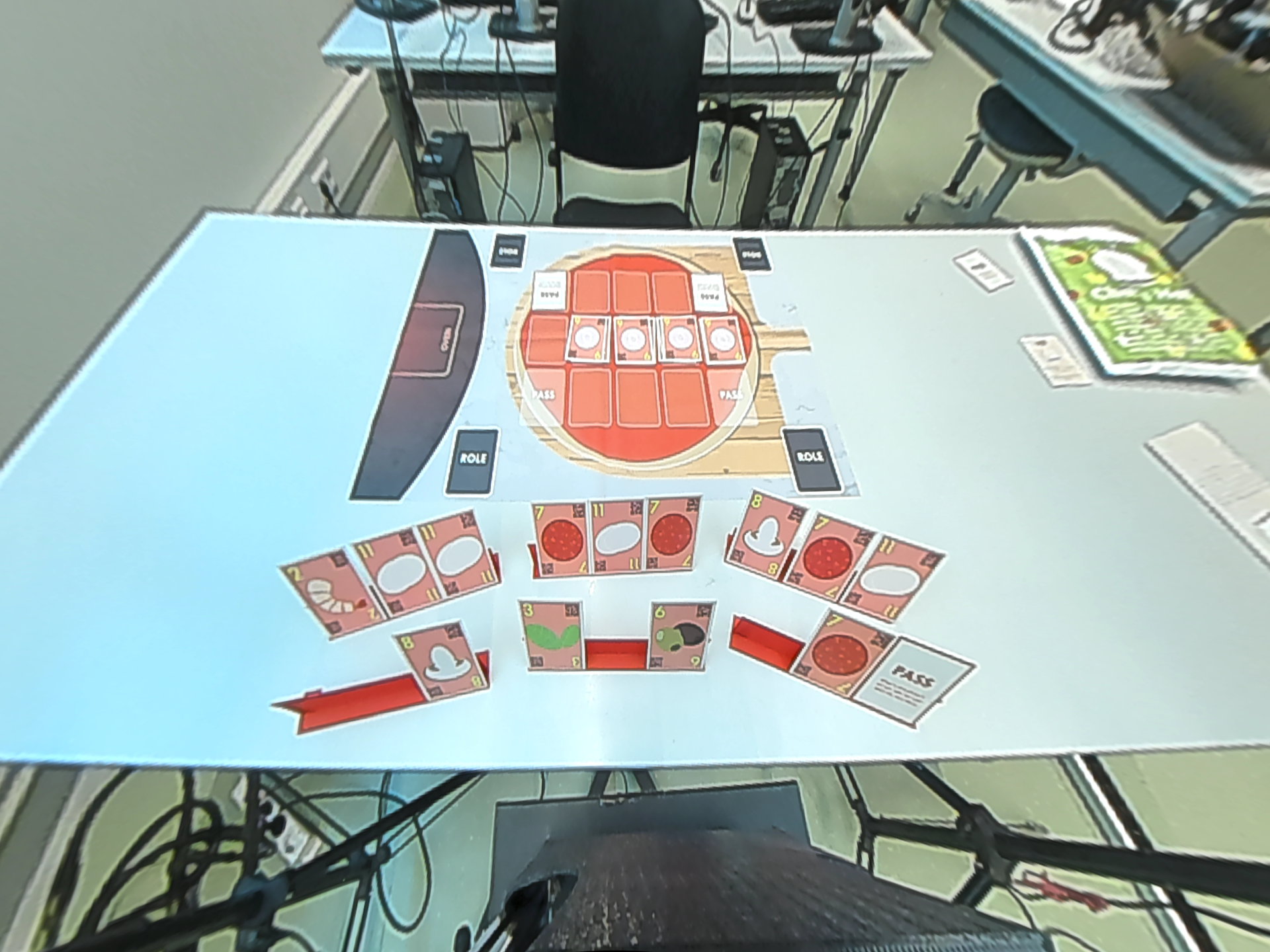}
  \caption{}
  \label{fig:real_data_plausible}
\end{subfigure}\hfill
\begin{subfigure}[t]{0.195\textwidth}
  \centering
  \includegraphics[width=\linewidth]{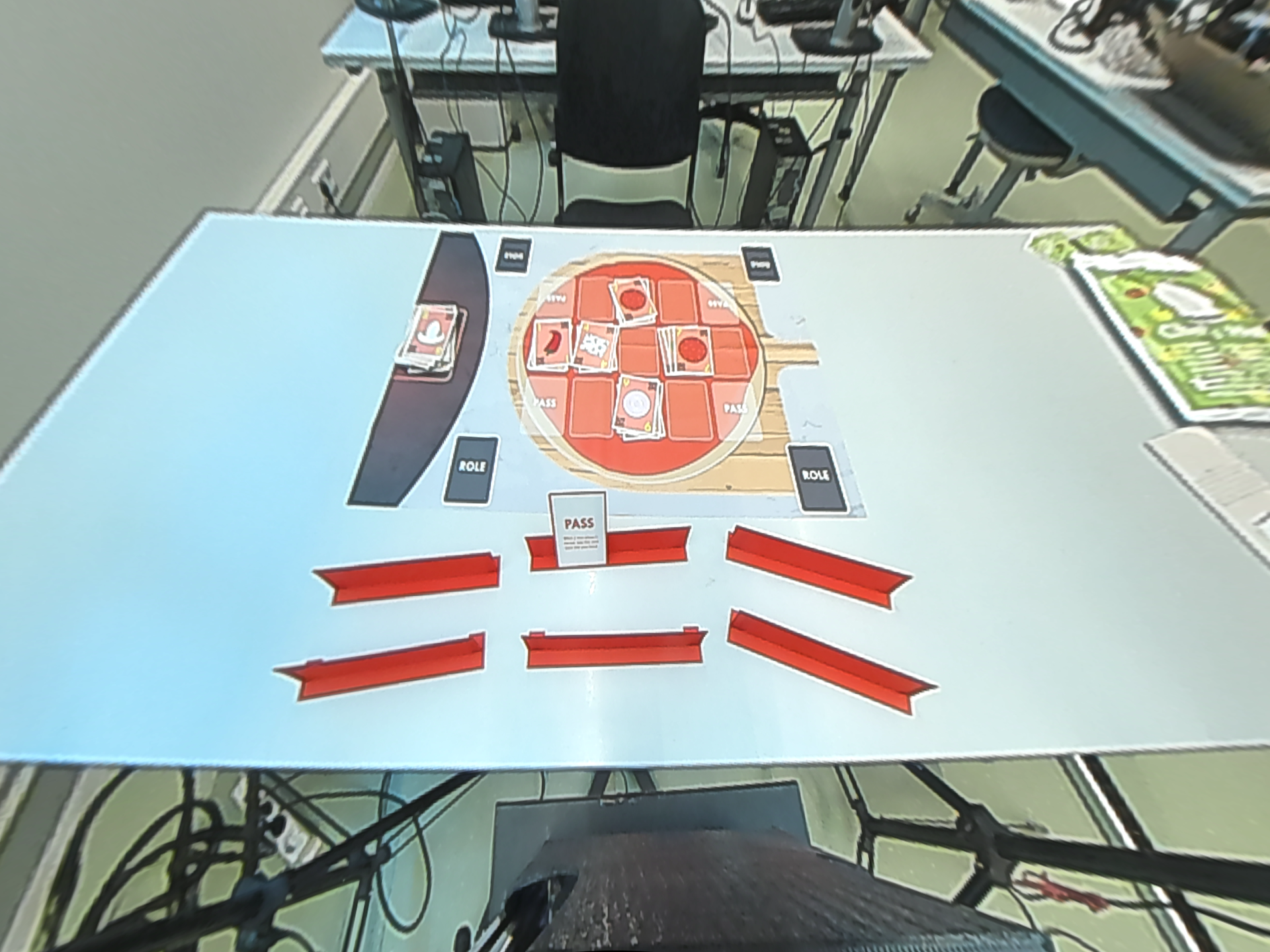}
  \caption{}
  \label{fig:real_data_stacked}
\end{subfigure}\hfill
\begin{subfigure}[t]{0.195\textwidth}
  \centering
  \includegraphics[width=\linewidth]{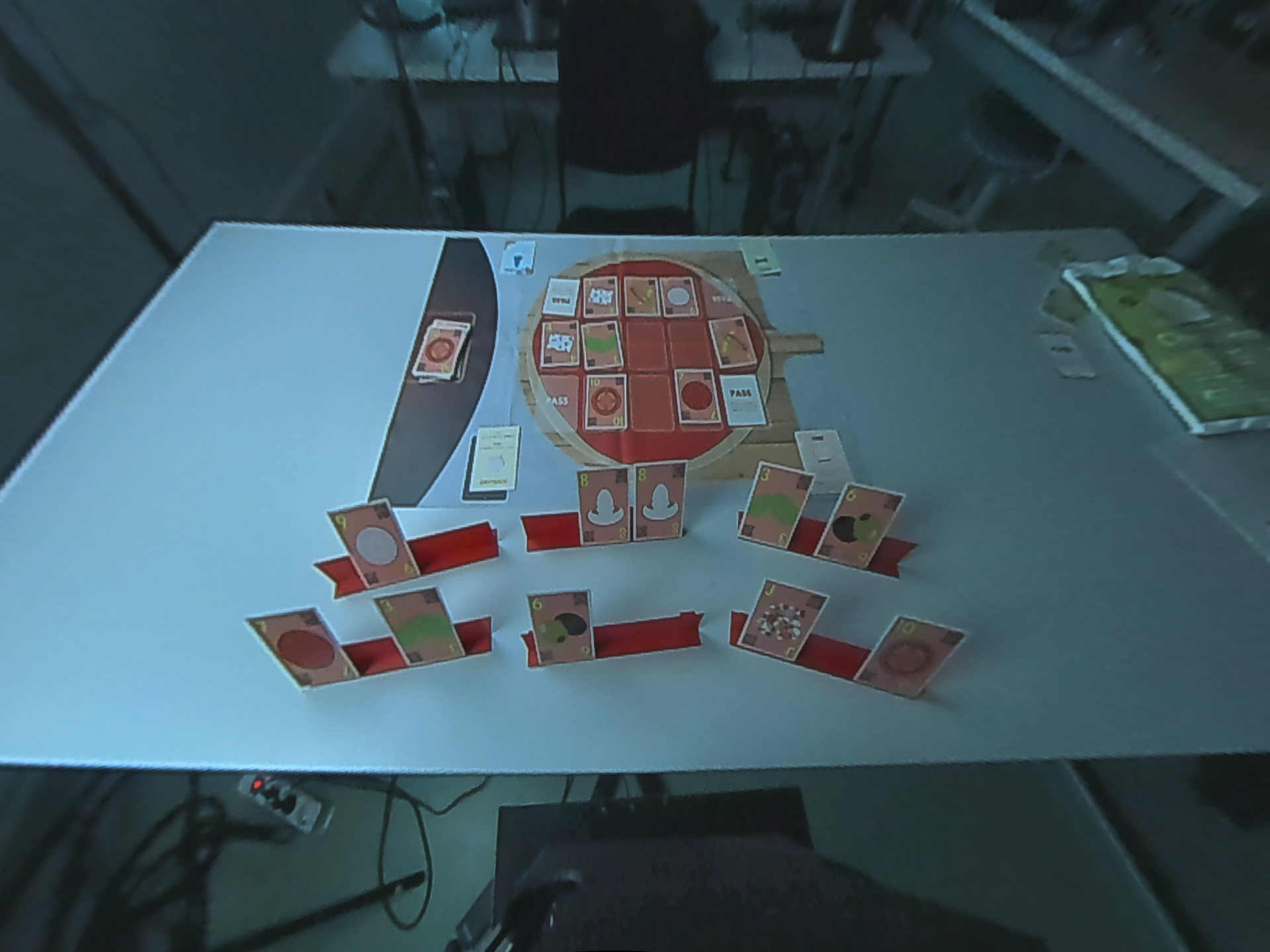}
  \caption{}
  \label{fig:real_data_directional}
\end{subfigure}\hfill
\begin{subfigure}[t]{0.195\textwidth}
  \centering
  \includegraphics[width=\linewidth]{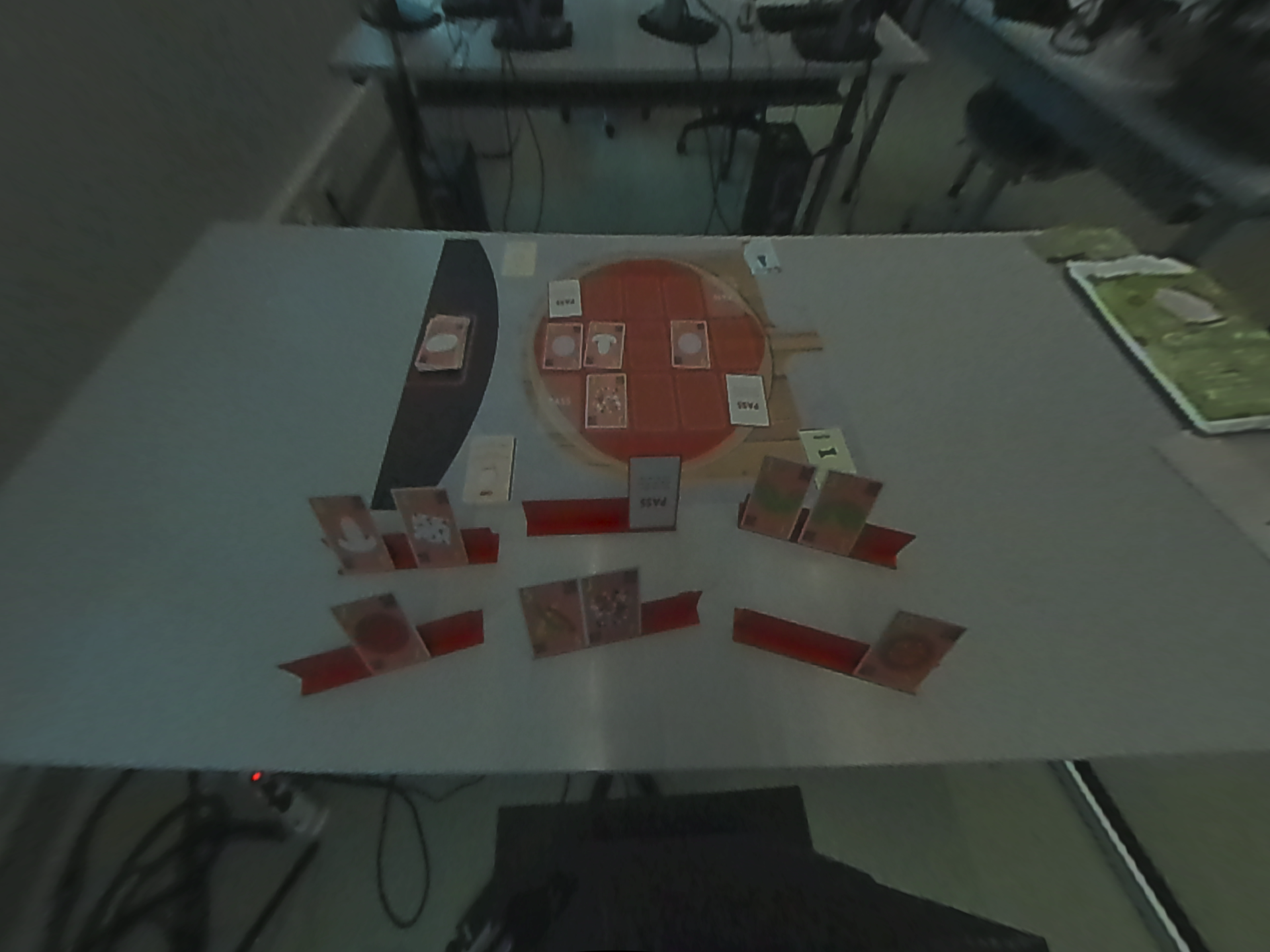}
  \caption{}
  \label{fig:real_data_dark}
\end{subfigure}
\caption{Real-world test images recorded under 5 conditions: random cards (a), plausible states (b), stacked cards (c), directional light (d), and low light (e)}
\label{fig:real_data}
\end{figure}

Our experiments evaluate the zero-shot sim-to-real transfer of our approach on a real-world test set of 100 rectified images recorded from the left fisheye camera of a real NICO robot under 5 conditions varying in lighting and card placement (see Figure \ref{fig:real_data}), without additional training. 

We evaluate both the standalone performance of the object detection model against human-labeled bounding boxes and the ability of the full pipeline to accurately infer the game state. %
For the latter, we count the number of cards of each class whose predicted bounding box center lies within the predicted segmentation mask of each game zone and compare them to the corresponding counts of the actual game state. Cards incorrectly assigned to a zone are false positives, while missing cards present in the actual game state are false negatives, allowing us to determine precision, recall, and the F1-score. %

Our evaluation analyzes the influence of our domain randomization approach by comparing the sim-to-real transfer of our finetuned object detection and segmentation models to YOLO26s and YOLO26s-seg models finetuned only on our canonical synthetic data using an 80/20 train-validation split. Additionally, we account for the influence of model size and variant by comparing the real-world performance of the \emph{n} and \emph{s} versions of both YOLO11 and YOLO26, each finetuned with our synthetic data.

\section{Results}

\begin{figure}[tp]
    \centering
    \begin{subfigure}{0.49\textwidth}
        \includegraphics[width=\textwidth]{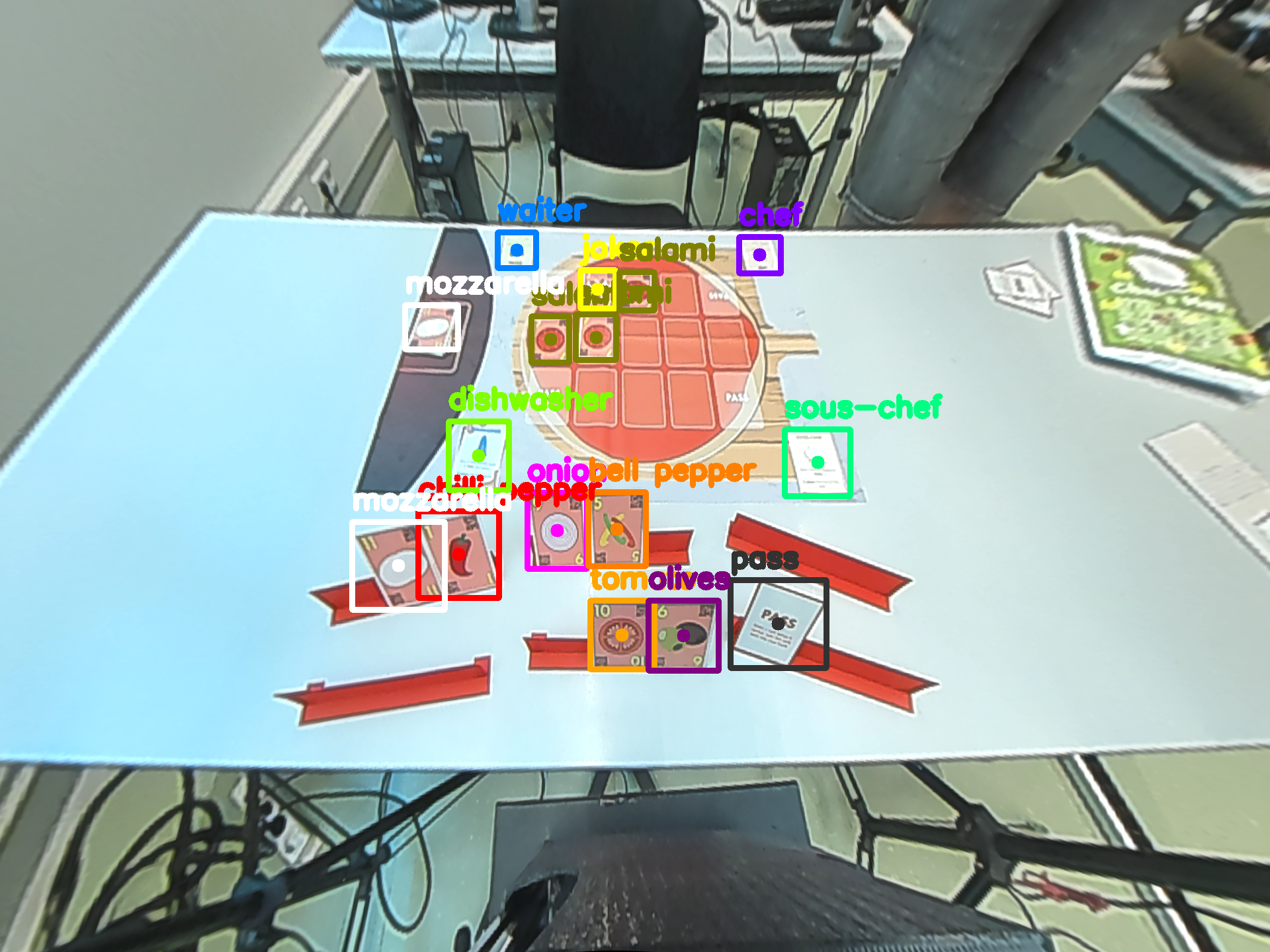}
        \caption{}
    \end{subfigure}
    \hfill
    \begin{subfigure}{0.49\textwidth}
        \includegraphics[width=\textwidth]{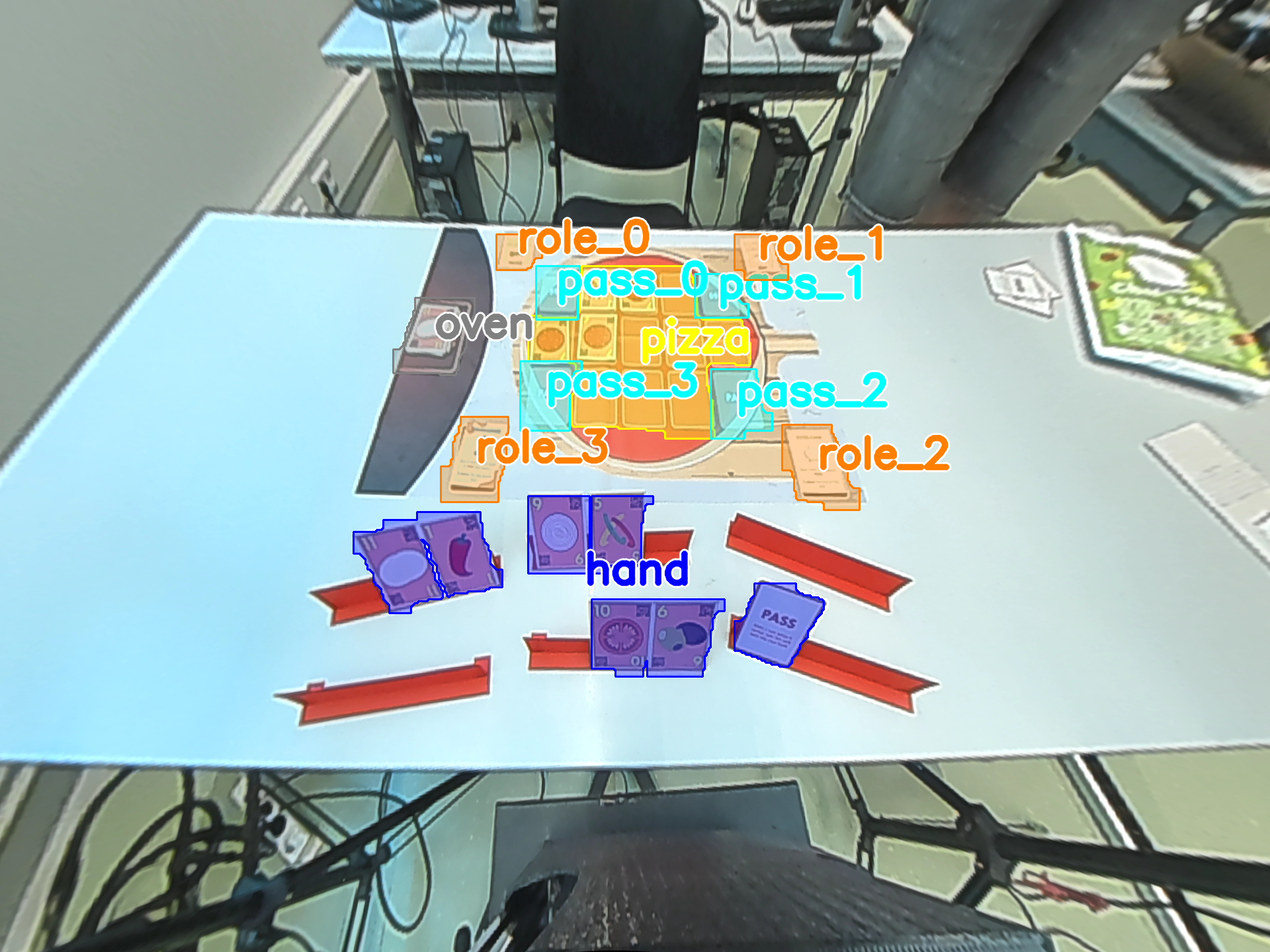}
        \caption{}
    \end{subfigure}
    \caption{Bounding boxes (a) and segmentation (b) on a real-world sample.}
    \label{fig_real_inference}
\end{figure}

In this section, we first report the standalone sim-to-real performance of our finetuned object detection model to locate and classify individual cards. Then, we report the real-world performance of our combined object detection and segmentation pipeline to accurately assign the detected cards to the different game zones and infer the game state.  Qualitative results for the real-world inference of the object detection and segmentation models are shown in Figure \ref{fig_real_inference}.

\subsection{Object Detection Performance}

\begin{table}[t!]
\centering
\caption{Real-world card detection performance of different YOLO object detection models finetuned exclusively with our synthetic data.}
\label{tab_result_detect}
\vspace{5pt}
\begin{tabular}{|l|c|c|c|c|c|c|}
\hline
Model & Infer.\ (ms) & P & R & F1 & mAP\textsubscript{50} & mAP\textsubscript{50:95}\\
\hline
YOLO26n            & \textbf{3.5} & 0.878 & 0.815 & \textbf{0.845} & \textbf{0.850} & \textbf{0.657}\\
YOLO26s            & 7.8 & 0.862 & \textbf{0.822} & 0.842 & 0.845 & 0.652\\
YOLO26s$^\dagger$  & 8.7 & 0.606 & 0.513 & 0.556 & 0.514 & 0.382\\
YOLO11n            & \textbf{3.5} & 0.845 & 0.781 & 0.812 & 0.811 & 0.619\\
YOLO11s            & 4.7 & \textbf{0.879} & 0.802 & 0.839 & 0.838 & 0.647\\
\hline
\multicolumn{7}{l}{$^\dagger$ Finetuned with canonical data only.}\\
\end{tabular}
\end{table}

As shown in Table \ref{tab_result_detect}, all four evaluated object detection models finetuned with our domain randomization approach significantly outperform the model finetuned with canonical data only. Overall, YOLO26n achieved the best object detection results with an F1-score of 0.845, an mAP\textsubscript{50} of 0.850, and an mAP\textsubscript{50:95} of 0.657, outperforming the YOLO26s finetuned exclusively with canonical data by 0.289, 0.336, and 0.275, respectively. The YOLO26 models generally outperform the YOLO11 models, although the difference is much more evident with YOLO11n, while YOLO11s achieves comparable results. Surprisingly, the increase in parameters between YOLO26n and YOLO26s %
yields worse results.

\subsection{Combined detection and segmentation pipeline}

\begin{table}[t!]
\centering
\caption{Real-world performance of our combined game state estimation pipeline trained exclusively on our synthetic data. To determine precision (P), recall (R), and F1-score (F1), we evaluate whether the right number of cards with the correct label are assigned to each zone. The missing avg reports the mean of ground-truth cards not assigned to a zone.}
\label{tab_result_combined}
\vspace{5pt}
\begin{tabular}{|l l|c|c|c|c|}
\hline
Models & & P & R & F1 & missing avg \\
\hline
YOLO26n & \& YOLO26n-seg  & \textbf{0.902} & 0.911 & \textbf{0.906} & 1.52\\
YOLO26s & \& YOLO26s-seg           & 0.899 & 0.906 & 0.902 & 1.61\\
YOLO26s$^\dagger$ & \& YOLO26s-seg$^\dagger$ & 0.727 & 0.551 & 0.627 & 7.02\\
YOLO11n & \& YOLO11n-seg & 0.835 & 0.908 & 0.870 & 1.38\\
YOLO11s & \& YOLO11s-seg & 0.873 & \textbf{0.918} & 0.895 & \textbf{1.34}\\
\hline
\multicolumn{5}{l}{$^\dagger$ Both models finetuned with canonical data only.}\\
\end{tabular}
\end{table}

The sim-to-real evaluation of our combined object detection and segmentation pipeline yields similar results, as shown in Table \ref{tab_result_combined}. Again, all models finetuned with our randomized data significantly outperform the models finetuned with canonical data only. The combination of YOLO26n and YOLO26n-seg achieved the highest F1-score of 0.906, missing 1.52 cards on average, whereas the pipeline finetuned exclusively with canonical data reached an F1-score of 0.627, with an average of 7.02 missing cards. As in the object detection case, the YOLO26 models achieved a higher F1-score than the YOLO11 models, with YOLO11s reaching results close to YOLO26s and the overall lowest average of 1.34 missing cards, albeit with a lower precision, indicating more false positives.

Thus, our results show that both for the detection of cards and for the segmentation of game zones to assign those cards to, our domain randomization approach greatly increases the zero-shot sim-to-real performance of a model finetuned exclusively on synthetic data, with YOLO26n and YOLO26n-seg reaching the best overall performance among the models evaluated.

\subsection{Discussion}
The ultimate goal of the presented research is to enable a robotic agent to autonomously play alongside human players in a game of Chef's Hat. Thus, it is important that the proposed game state detector is sufficient for this task. This sufficiency has two dimensions: speed and accuracy. As YOLO-architectures use a single forward pass to detect objects, the approach is real-time capable. Moreover, the fast processing would also allow capturing the game state from multiple viewpoints, e.g., head shifts, and over repeated observations.
This contributes to the second criterion, accuracy. The misdetection rate of approximately 1.5 cards implies that a single observation may not provide a perfect representation of the game state. However, human players are also unlikely to capture the complete game state with a single glance. Instead, they continuously update their internal representation of the game while observing other players' actions. 
Similarly, a robotic player can focus on the few cards played by other players at a time and incrementally update the game state. Inconsistencies with the known game state, the game logic, or assumptions about other players' goals and actions can further reveal detection errors.
This can be considered an application of the frame axiom: The game state can only change through player actions permitted by the game rules during their turn. We hypothesize that exploiting these constraints can lead to a sufficiently accurate game state representation in a real-world scenario without a perfect detection rate. %

\section{Conclusion}

We presented a fully automated synthetic data collection pipeline in CoppeliaSim for the Chef's Hat card game, which automatically generates labels through re-projection and applies domain randomization to bridge the reality gap, creating a dataset of 25413 samples without the need for manual annotation. Using this data, we finetuned a YOLO26n object detection model to locate and classify individual cards and a YOLO26n-seg segmentation model to identify game-relevant zones of the playing field, combining the two into a pipeline that infers the game state by determining which cards are placed in which zone.

Our results demonstrate that domain randomization is essential for the zero-shot sim-to-real transfer of our detection pipeline. The baseline models finetuned exclusively on canonical synthetic data are unable to fully assess the game state with an F1-score of 0.627, missing 7.02 cards on average. In contrast, our domain-randomized models reduce the missing card average to 1.52 and achieve an F1-score of 0.906. Considering the intentionally adverse lighting conditions in some of our test cases, this demonstrates the resilience of our model.

While our pipeline achieves strong card-level detection, fully autonomous gameplay intuitively requires perfect game state inference. Future work could address this directly by incorporating more realistic camera simulation and extending the dataset with card back sides, distractor objects, and more sophisticated lighting. However, as discussed, we hypothesize that exploiting game logic constraints allows an agent to act reliably under imperfect perception, which we will explore in future work. Ultimately, the goal is to integrate our perception pipeline with an embodied agent such as ELMiRA \cite{gade2024embodying} to enable fully autonomous play of Chef's Hat, combining game state inference with physical card manipulation and social interaction with human players.

\begin{credits}
\subsubsection{\ackname} The authors gratefully acknowledge support from the German Research Foundation DFG under project LUMO (No 551629603) and funding from Horizon Europe under the MSCA grant agreements No 101226624 (GREET), No 101168792 (SWEET), and No 101072488 (TRAIL).

\subsubsection{\discintname}
The authors have no competing interests to declare that are relevant to the content of this article. 
\end{credits}
\bibliographystyle{splncs04}
\bibliography{references_clean}
\end{document}